\documentclass[10pt, letter, twocolumn]{article} 

\usepackage[english]{babel} 

\usepackage{microtype} 

\usepackage{amsmath,amsfonts,amsthm} 

\usepackage[svgnames]{xcolor} 
\usepackage{mathtools}
\usepackage{array}
\usepackage{hyperref}

\usepackage{color}

\usepackage{flushend}

\definecolor{boxblue}{RGB}{234,249,254}
\definecolor{lineblue}{RGB}{141,214,239}

\definecolor{light}{rgb}{0.4, 0.4, 0.4}
\definecolor{lblue}{rgb}{0.56,0.80,0.92}
\definecolor{llblue}{rgb}{0.9, 0.9, 0.9}
\definecolor{dgrey}{rgb}{0.2, 0.2, 0.2}
\definecolor{lgrey}{rgb}{0.5, 0.5, 0.5}
\def\light#1{{\color{light}#1}}

\usepackage[footnotesize, up, textfont=sf]{caption} 

\DeclareCaptionFormat{figfont}{{\color{lblue}\usefont{OT1}{ugq}{n}{n}{#1. \ }}#3}
\DeclareCaptionFormat{tabfont}{{\color{lblue}\usefont{OT1}{ugq}{n}{n}{#1. \ }}#3}
\usepackage{booktabs} 
\usepackage{lastpage} 

\usepackage{graphicx,dblfloatfix} 
\usepackage{tikz}
\usetikzlibrary{backgrounds,calc}
\usepackage{enumitem} 
\usepackage[framemethod=TikZ]{mdframed}

\newsavebox{\picbox}

\setlist{noitemsep} 
\usepackage{todonotes}

\usepackage{sectsty} 
\sectionfont{
    \Large\color{lblue}\usefont{OT1}{ugq}{n}{n}
}

\subsectionfont{
    \color{light}\usefont{OT1}{ugq}{n}{n}
}

\hypersetup{
    colorlinks=true,
    linkcolor=black,
    urlcolor=black,
    citecolor=black
}

\usepackage{geometry} 

\usepackage[T1]{fontenc} 
\usepackage[utf8]{inputenc} 

\usepackage{anyfontsize}

\usepackage{fancyhdr} 
\fancypagestyle{firstpage}{ 
	\fancyhf{}
    \lfoot{
        \scriptsize
        \light{
            {\fontfamily{cmss}\selectfont
            This article has been accepted for publication in IEEE Robotics \& Automation Magazine \\ Published Version Available at \url{https://doi.org/10.1109/MRA.2023.3270228}
            }
        }
    }
    \rfoot{\footnotesize
    \fontfamily{cmss}\selectfont
    \light{\thepage}\
    } 
}

\usepackage[firstpage=true]{background}

\backgroundsetup{
scale=1,
angle=0,
opacity=1,
contents={\begin{tikzpicture}[remember picture,overlay]
 \path [top color = white, bottom color = llblue] ($(current page.north west)-(4in,4in)$) rectangle (current page.south east);
\end{tikzpicture}}
}

\newcommand{\authorstyle}[1]{\color{dgrey}\centering{\usefont{OT1}{phv}{n}{n}#1}} 

\newcommand{\titlestyle}[1]{\color{dgrey}\fontsize{35}{45}\centering{\usefont{OT1}{ugq}{n}{n}#1}}

\usepackage{titling} 

\pretitle{
	\vspace{110pt} 
	\fontsize{40}{46}\usefont{OT1}{phv}{b}{n}\selectfont 
}

\posttitle{\par\vskip 10pt} 

\preauthor{} 

\postauthor{ 
	\vspace{-15pt} 
}

\usepackage{lettrine} 
\newcommand{\initial}[1]{ 
	\lettrine[lines=4,findent=5pt,nindent=0pt]{
		\color{lblue}
		{#1}
	}{}%
}

\usepackage[backend=biber,style=ieee,natbib=true,doi=false,isbn=false,url=false,eprint=false]{biblatex} 

\AtNextBibliography{\footnotesize}
\usepackage[autostyle=true]{csquotes} 

\usepackage{wrapfig}

\usepackage{bbding}

\usepackage{amsmath}
\usepackage{amssymb}
\usepackage{accents}
\usepackage{bm}
\usepackage{xifthen}
\usepackage{color}
\usepackage{fancyvrb}
\usepackage{graphicx,scalerel}

\newcommand{\ba}{\begin{eqnarray}}
\newcommand{\ea}{\end{eqnarray}}
\newcommand{\eq}[1]{(\ref{#1})}

\newcommand{\R}{\mathbb{R}}

\newcommand{\presup}[1]{\,{}^{\scriptscriptstyle #1}\!}

\newcommand{\pose}[1][_NONE]{\ifthenelse{\equal{#1}{_NONE}}{}{\presup{#1}}{\mathbf{\xi}}}
\newcommand{\estpose}[1][_NONE]{\ifthenelse{\equal{#1}{_NONE}}{}{\presup{#1}}{\hat{\mathbf{\xi}}}}
\newcommand{\hpose}[1][_NONE]{\ifthenelse{\equal{#1}{_NONE}}{}{\presup{#1}}{\hat{\mathbf{\xi}}}}
\newcommand{\posedot}[1][_NONE]{\ifthenelse{\equal{#1}{_NONE}}{}{\presup{#1}}{\mathbf{\nu}}}
\newcommand{\q}[1][_NONE]{\ifthenelse{\equal{#1}{_NONE}}{}{\presup{#1}}{\mathring{q}}}
\newcommand{\uquat}[2][_NONE]{\ifthenelse{\equal{#1}{_NONE}}{}{\presup{#1}}{\mathring{#2}}}
\newcommand{\duquat}[2][_NONE]{\ifthenelse{\equal{#1}{_NONE}}{}{\presup{#1}}{\dot{\mathring{#2}}}}
\newcommand{\quat}[2][_NONE]{\ifthenelse{\equal{#1}{_NONE}}{}{\presup{#1}}{\breve{#2}}}
\DeclareMathAlphabet{\mathitbf}{OML}{cmm}{b}{it}
\newcommand{\twist}[2][_NONE]{\ifthenelse{\equal{#1}{_NONE}}{}{\presup{#1}}{\mathcal{S}}}
\renewcommand{\vec}[2][_NONE]{\ifthenelse{\equal{#1}{_NONE}}{}{\presup{#1}}{\boldsymbol #2}}
\newcommand{\hvec}[2][_NONE]{\ifthenelse{\equal{#1}{_NONE}}{}{\presup{#1}}{\tilde{\vec{#2}}}}
\newcommand{\evec}[2][_NONE]{\ifthenelse{\equal{#1}{_NONE}}{}{\presup{#1}}{\hat{\vec{#2}}}}
\newcommand{\bvec}[2][_NONE]{\ifthenelse{\equal{#1}{_NONE}}{}{\presup{#1}}{\bar{\vec{#2}}}}
\newcommand{\dhvec}[2][_NONE]{\ifthenelse{\equal{#1}{_NONE}}{}{\presup{#1}}{\dot{\tilde{\vec{#2}}}}}
\newcommand{\dvec}[2][_NONE]{\ifthenelse{\equal{#1}{_NONE}}{}{\presup{#1}}{\dot{\vec{#2}}}}
\newcommand{\ddvec}[2][_NONE]{\ifthenelse{\equal{#1}{_NONE}}{}{\presup{#1}}{\ddot{\vec{#2}}}}

\newcommand{\mat}[2][_NONE]{\ifthenelse{\equal{#1}{_NONE}}{}{\presup{#1}\,}{{\mathbf #2}}}
\newcommand{\dmat}[2][_NONE]{\ifthenelse{\equal{#1}{_NONE}}{}{\presup{#1}\,}{{\dot{\mathbf #2}}}}
\newcommand{\emat}[2][_NONE]{\ifthenelse{\equal{#1}{_NONE}}{}{\presup{#1}\,}{\hat{\mathbf#2}}}
\newcommand{\matfn}[3][_NONE]{\ifthenelse{\equal{#1}{_NONE}}{}{\presup{#1}}{{\mat{#2}}\left(#3\right)}}
\newcommand{\Rt}[2][_NONE]{\ifthenelse{\equal{#1}{_NONE}}{}{\presup{#1}}{{\bf R}\left(#2\right)}}
\newcommand{\cframe}[1]{\ensuremath{\{\mathrm{#1}\}}}

\newcommand{\point}[2][_NONE]{\ifthenelse{\equal{#1}{_NONE}}{}{\ensuremath{\presup{#1}}}{\ensuremath{\mathrm{#2}}}}

\newfont{\School}{pncr}
\newfont{\eightTR}{pncr at 8pt}

\fvset{formatcom={\color{blue}\baselineskip=-\maxdimen\lineskip=0pt},xleftmargin=4mm,commentchar=!}
\DefineVerbatimEnvironment{Code}{Verbatim}{}
\DefineVerbatimEnvironment{CodeSmall}{Verbatim}{fontsize=\scriptsize,xleftmargin=1mm,commentchar=!}
\DefineVerbatimEnvironment{CodeNum}{Verbatim}{numbers=left,numbersep=4pt,fontsize=\footnotesize,xleftmargin=4mm}
\newcommand{\func}[2][_NONE]{\ifthenelse{\equal{#1}{_NONE}}{\index{code}{#2}}{\index{code}{#1}}\ifthenelse{\boolean{draft}}{{\color{green}\Verb+#2+}}{\Verb+#2+}}
\newcommand{\methodb}[2]{\index{code}{#1@\textbf{#1}!.#2}\ifthenelse{\boolean{draft}}{{\color{magenta}\Verb+#1.#2+}}{\Verb+#1.#2+}}
\newcommand{\method}[2]{\index{code}{#1@\textbf{#1}!.#2}\ifthenelse{\boolean{draft}}{{\color{magenta}\Verb+#2+}}{\Verb+#2+}}
\newcommand{\class}[1]{\index{code}{#1@\textbf{#1}}\ifthenelse{\boolean{draft}}{{\color{cyan}\Verb+#1+}}{\Verb+#1+}}
\newcommand{\property}[1]{\index{property}{#1}\ifthenelse{\boolean{draft}}{{\color{cyan}\Verb+#1+}}{\Verb+#1+}}

\newcommand{\SE}[1]{\ensuremath{\mathrm{{\bf SE}(#1)}}}
\newcommand{\SO}[1]{\ensuremath{\mathrm{{\bf SO}(#1)}}}

\newcommand{\isk}[1]{\vee\left( #1\right)}
\newcommand{\iskx}[1]{\vee_{\times}\left( #1\right)}
\newcommand{\skx}[1]{\left[#1\right]_{\times}}
\newcommand{\sk}[1]{\left[#1\right]}

\usepackage{xparse}
\NewDocumentCommand{\ovec}{ o m o }{%
  \IfValueT{#1}{\presup{#1}}%
  \vec{#2}%
  \IfValueT{#3}{_{#3}}
  ^{\scalebox{0.6}{\#}}%
  }
\NewDocumentCommand{\omat}{ o m o }{%
  \IfValueT{#1}{\presup{#1}}%
  \mat{#2}%
  \IfValueT{#3}{_{#3}}
  ^{\scalebox{0.6}{\#}}%
  }
\NewDocumentCommand{\obspose}{ o o }{%
  \IfValueT{#1}{\presup{#1}}%
  \mathbf{\xi}%
  \IfValueT{#2}{_{#2}}
  ^{\scalebox{0.6}{\#}}%
  }

\renewcommand{\q}{\vec{q}}
\newcommand{\w}{\boldsymbol{\omega}}

\newcommand{\qd}{\dot{\vec{q}}}

\newcommand{\J}{\mat{J}}

\newcommand{\Jt}{\mat{J_{\!t}}}
\newcommand{\JR}{\mat{J_{\!R}}}
\renewcommand{\R}{\mat{R}}
\newcommand{\Rd}{\dot{\mat{R}}}

\newcommand{\T}{\mat{T}}
\newcommand{\Td}{\dot{\mat{T}}}

\renewcommand{\t}{\vec{t}}
\newcommand{\td}{\dot{\vec{t}}}

\newcommand{\Jv}{\mat{J}_v}
\newcommand{\Jvx}[1]{\mat{J}_{{\!\nu}_{#1}}}
\newcommand{\Jw}{\mat{J_{\!\omega}}}
\newcommand{\Jwx}[1]{\mat{J}_{{\!\omega}_{#1}}}

\newcommand{\TR}[1]{\mat{T}_{{\!\R}_{#1}}}
\newcommand{\Tt}[1]{\mat{T}_{{\!\t}_{#1}}}

\usepackage{newfloat}
\DeclareFloatingEnvironment[
    fileext=los,
    listname={List of Excurses},
    name=Excurse,
    placement=tbhp,
    within=section,
]{excurse}

\DeclareCaptionFormat{excfont}{
    \normalsize{
        \color{lblue}
        \usefont{OT1}{ugq}{n}{n}{
            #1. \
        }
        \color{dgrey} #3
    }
}
\title{
\vspace{-148pt}
\protect\centering \protect\includegraphics[width=0.35\linewidth]{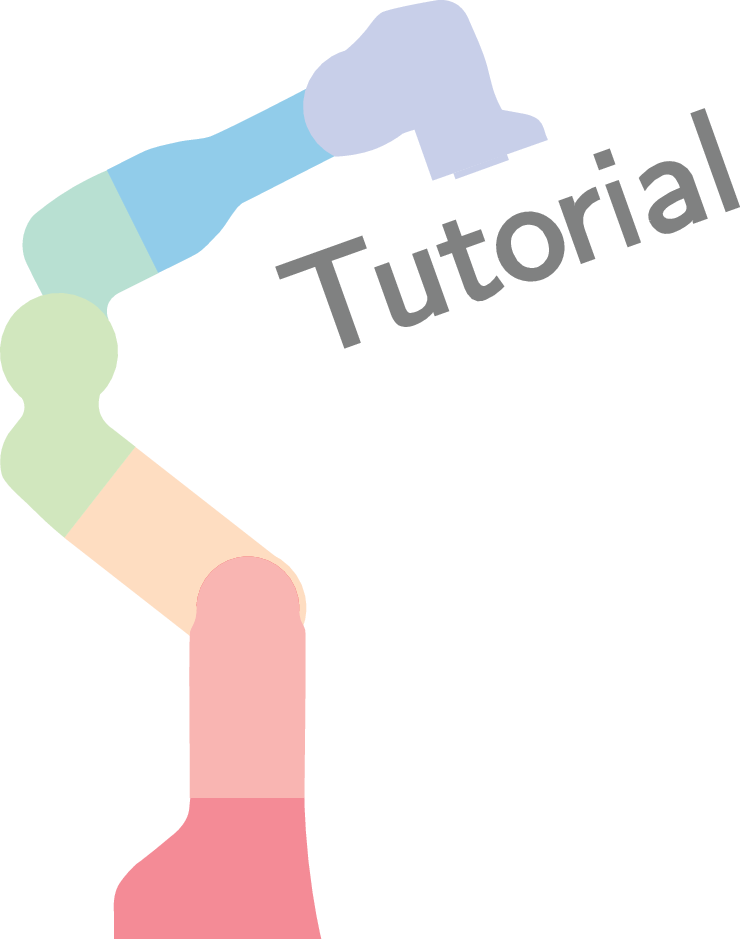}\\
\vspace{32pt}
\Huge
\titlestyle{Manipulator \\\vspace{5pt}Differential Kinematics}\\
\vspace{10pt}
\protect\centering\Large Part 1: Kinematics, Velocity, and Applications\\
[0.2cm]
}

\author{
    {
       \authorstyle{By Jesse Haviland and Peter Corke}\\[0.5cm]
    }
}

\date{}

\begin{document}

\maketitle
\thispagestyle{firstpage}

\initial{K}inematics, derived from the Greek word for motion, is the branch of mechanics that studies the motion of a body, or a system of bodies, without considering mass or force. This two-part tutorial is about the kinematics of robot 
manipulators, and in that context, it is concerned with the relationship
between the position of the robot's joints and the pose of its end effector, as well as the relationships between various derivatives of those quantities.
Kinematics is a fundamental concept in the study or application of robot manipulators, and our audience for Part 1 is students, practitioners or researchers encountering this topic for the first time, or looking for a concise refresher.

The relationship between end-effector pose and joint coordinates is encapsulated in the forward and inverse kinematics, and these are amongst the first concepts learned in this field.  This tutorial begins with forward kinematics, but deliberately avoids the commonly-used Denavit-Hartenberg parameters which we feel confound pedagogy. Instead, we approach the problem using the elementary transform sequence (ETS) which is an intuitive, uncomplicated, and superior method for modelling a kinematic chain. Then we formulate the first-order differential kinematics which is the relationship between the velocity of the robot's joints and the end-effector velocity. This relationship is expressed in terms of the manipulator Jacobian and is foundational for many fundamental robotic control algorithms, three of which we cover in this first article. Firstly, resolved-rate motion control -- a simple algorithm that enables reactive, closed-loop and sensor-based velocity control of a manipulator. Secondly, numerical inverse kinematics (IK); an important planning tool that solves for joint coordinates which correspond to a specific manipulator pose. We introduce various methods that combine first-order differential kinematics and numerical optimisation and present a comprehensive experiment that compares the performance and characteristics of each numerical IK method. Thirdly, we introduce performance measures that describe a manipulator's ability to move or exert force depending on its joint configuration.  

\begin{figure}[!b]
    \centering
    \includegraphics[height=14.5cm]{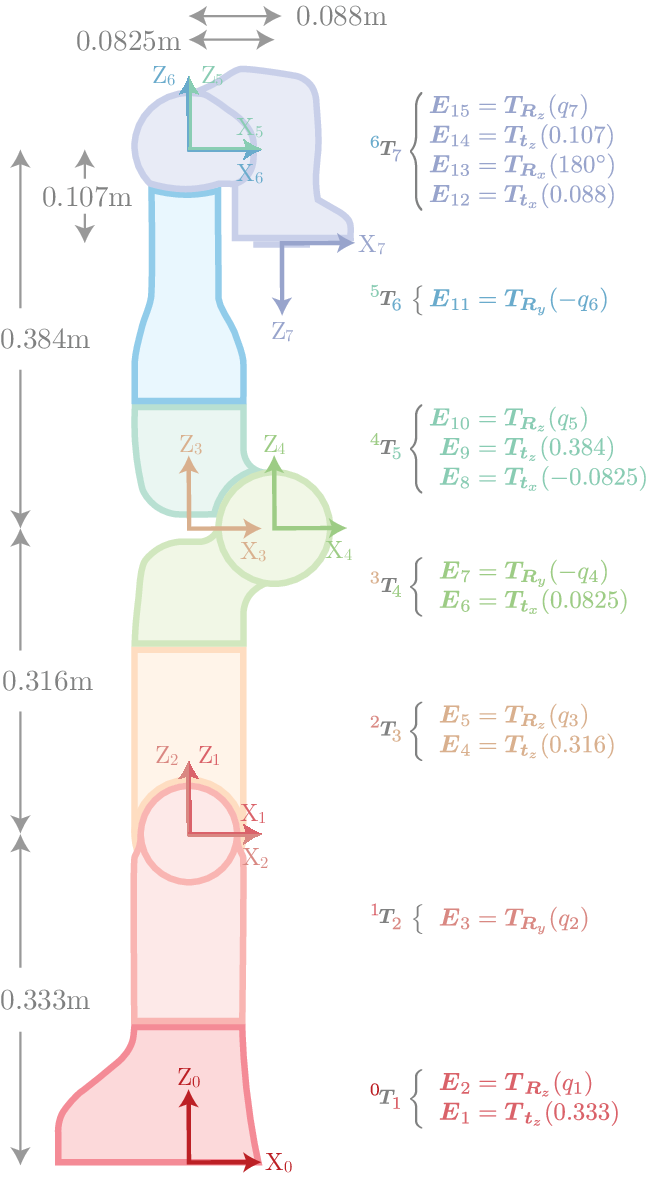}
    \caption{
        The Elementary Transform Sequence of the seven degree-of-freedom Franka-Emika Panda serial-link manipulator in its zero-angle configurations. $\mat{E}_i$ represents an elementary transform while $^a\mat{T}_b$ 
        represents the pose of link frame $b$ in the reference frame of link $a$.
    }
    \label{fig:cover}
\end{figure}

\begin{figure}[!b]
    \centering
    \includegraphics[height=14.5cm]{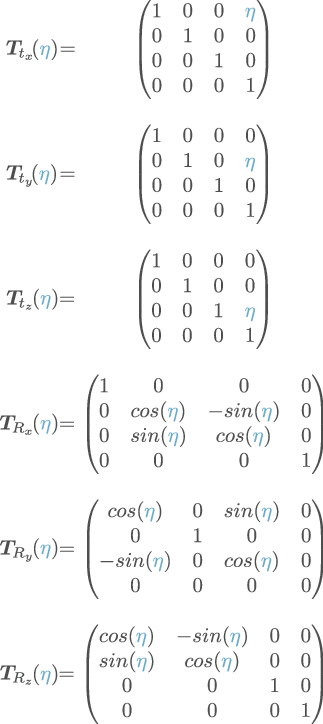}
    \caption{
        The six elementary transforms $\mat{E} \in \SE{3}$ from (\ref{eq:Ei}) which are the building blocks for ETS notation. Each homogeneous transformation matrix represents a translation along, or a rotation about, a single axis which is parameterized by $\eta$ as defined in (\ref{eq:eta1}) and (\ref{eq:eta2})
    }
    \label{fig:et}
\end{figure}

Part 2 \cite{dkt2} of this tutorial provides a formulation of second and higher-order differential kinematics, introduces the manipulator Hessian, and illustrates advanced techniques, many of which improve the performance of techniques demonstrated in Part 1. These are useful topics that are not well covered in current textbooks.

We have provided Jupyter Notebooks to accompany this tutorial. The Notebooks are written in Python and use the Robotics Toolbox for Python, and the Swift Simulator \cite{rtb} to provide full implementations of each concept, equation, and algorithm presented in this tutorial. The Notebooks use rich Markdown text and \LaTeX\ equations to document and communicate key concepts. While not absolutely essential, for the most engaging and informative experience, we recommend working through the Jupyter Notebooks while reading this article. The Notebooks and setup instructions can be accessed at \href{https://github.com/jhavl/dkt}{github.com/jhavl/dkt}.

A serial-link manipulator, which we refer to as a manipulator, is the formal name for a robot that comprises a chain of rigid links and joints, it may contain branches, but it cannot have closed loops. Each joint provides one degree of freedom, which may be a prismatic joint providing translational freedom or a revolute joint providing rotational freedom. The base frame of a manipulator represents the reference frame of the first link in the chain, while the last link is known as the end-effector.

The elementary transform sequence (ETS), introduced in \cite{ets}, provides a universal method for describing the kinematics of any manipulator. This intuitive and systematic approach can be applied with a simple \emph{walk through} procedure. The resulting sequence comprises a number of elementary transforms -- translations and rotations -- from the base frame to the robot's end-effector. An example of an ETS is displayed in Figure \ref{fig:cover} for the Franka-Emika Panda in its zero-angle configuration.

The ETS is conceptually easy to grasp, and a reader can intuitively understand the kinematics of a manipulator at a glance. Additionally, the ETS avoids unnecessary complexity and frame assignment constraints of Denavit and Hartenberg (DH) notation \cite{dh} and allows joint rotation or translation about or along any axis. This makes the ETS a powerful tool for understanding robot kinematics and going forward we believe the ETS method should be considered the standard method for describing robot kinematics.

We use the notation of \cite{peter} where $\cframe{a}$ denotes a coordinate frame, and $^a\mat{T}_b$ is a relative pose or rigid-body transformation of $\cframe{b}$ with respect to $\cframe{a}$.

\begin{figure*}[!b]
    \centering
    \includegraphics[height=4.5cm]{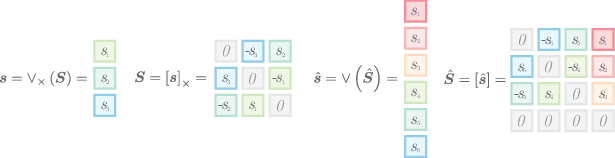}
    \caption{
        Shown on the left is a vector $\vec{s}\in \mathbb{R}^3$ along with its corresponding skew symmetric matrix $\mat{S} \in \bf{so}(3) \subset \mathbb{R}^{3 \times 3}$. Shown on the right is a vector $\evec{s}\in \mathbb{R}^6$ along with its corresponding augmented skew symmetric matrix $\emat{S} \in \bf{se}(3) \subset \mathbb{R}^{4 \times 4}$. The skew functions $\skx{\cdot} : \mathbb{R}^3 \mapsto \bf{so}(3)$ maps a vector to a skew symmetric matrix, and $\sk{\cdot} : \mathbb{R}^6 \mapsto \bf{se}(3)$ maps a vector to an augmented skew symmetric matrix. The inverse skew functions $\iskx{\cdot} : \bf{so}(3) \mapsto \mathbb{R}^3$ maps a skew symmetric matrix to a vector and $\isk{\cdot} : \bf{se}(3) \mapsto \mathbb{R}^6$ maps an augmented skew symmetric matrix to a vector.
    }
    \label{fig:skew}
\end{figure*}


\section*{Forward Kinematics}\label{sec:ets}

The forward kinematics is the first and most fundamental relationship between the link geometry and robot configuration.

The forward kinematics of a manipulator provides a non-linear mapping
\begin{equation*}
    {^0\T_e(t)} = {\cal K}(\q(t))
\end{equation*}
between the joint space and Cartesian task space,
where $\q(t) = (q_1(t), q_2(t), \cdots q_n(t)) \in \mathbb{R}^n$ is the vector of joint generalised coordinates, $n$ is the number of joints, and ${^0\T_e}  \in \SE{3}$ is a homogeneous transformation matrix representing the pose of the robot's end-effector in the world-coordinate frame. The ETS model defines $\cal{K}(\cdot)$ as the product of $M$ elementary transforms $\mat{E}_i \in \SE{3}$
\begin{align} \label{m:ets1}
    {^0\T_e(t)}  &= \mat{E}_1(\eta_1) \ \mat{E}_2(\eta_2) \ \hdots \ \mat{E}_M(\eta_M) \nonumber\\
    &= \prod_{i=1}^{M} \mat{E}_i(\eta_i).
\end{align}

Each of the elementary transforms $\mat{E}_i$ can be a pure translation along, or a pure rotation about the local x-, y-, or z-axis by an amount $\eta_i$. Explicitly, each transform is one of the following
\begin{equation} \label{eq:Ei}
    \mat{E}_i = 
    \left\{
    \begin{matrix}
        \Tt{x}(\eta_i) \\
        \Tt{y}(\eta_i) \\
        \Tt{z}(\eta_i) \\
        \TR{x}(\eta_i) \\
        \TR{y}(\eta_i) \\
        \TR{z}(\eta_i) \\
    \end{matrix}
    \right.
\end{equation}
where each of the matrices are displayed in Figure \ref{fig:et} and the parameter $\eta_i$ is either a constant $c_i$ (translational offset or rotation) or a joint variable $q_j(t)$
\begin{equation} \label{eq:eta1}
    \eta_i = 
    \left\{
    \begin{matrix}
        c_i \\
        q_j(t) \\
    \end{matrix}
    \right. 
\end{equation}
and the joint variable is
\begin{equation} \label{eq:eta2}
    q_j(t) = 
    \left\{
    \begin{matrix*}[l]
        \theta_j(t) & \quad \mbox{for a revolute joint}\\
        d_j(t) & \quad \mbox{for a prismatic joint}\\
    \end{matrix*}
    \right. 
\end{equation}
where $\theta_j(t)$ represents a joint angle, and $d_j(t)$ represents a joint translation.

An ETS description does not require intermediate link frames, but it does not preclude their introduction.  The convention we adopt is to place the $j^{th}$ frame $\{j\}$
immediately after the ETS term related to $q_j$, as shown in Figure \ref{fig:cover}.
The relative transform between link frames $a$ and $b$ is simply a subset of the ETS
\begin{align} \label{m:joint_pose}
    ^a\mat{T}_b &= \prod_{i=\mu(a)}^{\mu(b)} \mat{E}_i(\eta_i)
\end{align}
where the function $\mu(j)$ returns the index of the factor in the ETS expression, (\ref{m:ets1}) in this case, which is a function of $q_j$. For example, from Figure \ref{fig:cover}, for joint variable $j = 5$, $\mu(j) = 10$.


\section*{Deriving the Manipulator Jacobian}
\subsection*{First Derivative of a Pose}

In the following sections, we use the skew and inverse skew operation as defined in Figure \ref{fig:skew}. Now consider the end-effector pose, which is a function of the joint coordinates given by (\ref{m:ets1}). Its derivative with respect to time is
\begin{equation} \label{eq:Tdot}
\Td = 
    \frac{\mathrm{d} \mat{T}}
         {\mathrm{d}t} 
    = 
    \frac{\partial \T}
         {\partial q_1} \dot{q}_1 + \cdots +  
    \frac{\partial \T}
         {\partial q_n} \dot{q}_n \in \mathbb{R}^{4 \times 4} 
\end{equation}
where each $\frac{\partial \T}{\partial q_i} \in \mathbb{R}^{4 \times 4}$.

The information in $\T$ is non-minimal, and redundant, as is the information in $\Td$.  We can write these respectively as
\begin{equation}
     \T = 
     \begin{pmatrix}
          \R & \t \\ 
          0 & 1
     \end{pmatrix}
     , \,\,\, \Td = 
     \begin{pmatrix}
          \Rd & \td \\ 
          0 & 0 
     \end{pmatrix}
\end{equation}
where $\R \in \SO{3}$, $\Rd \in \mathbb{R}^{3 \times 3}$, and $\t, \td \in \mathbb{R}^3$.

We will write the partial derivative in partitioned form as
\begin{equation} \label{eq:partition}
     \frac{\partial \T}
          {\partial q_j} = 
     \begin{pmatrix}
          \JR_j & \Jt_j \\ 
          0 & 0 
     \end{pmatrix} 
\end{equation}
where $\JR_j \in \mathbb{R}^{3 \times 3}$ and $\Jt_j \in \mathbb{R}^{3 \times 1}$, and then rewrite \eq{eq:Tdot} as 
\begin{equation*}
     \begin{pmatrix}
          \Rd & \td \\ 0 & 0 
     \end{pmatrix} = 
     \begin{pmatrix} 
          \JR_1 & \Jt_1 \\ 0 & 0 
     \end{pmatrix}
     \dot{q}_1 + \cdots +   
     \begin{pmatrix} 
          \JR_n & \Jt_n \\ 0 & 0 
     \end{pmatrix}
     \dot{q}_n \,\,
\end{equation*}
and write a matrix equation for each non-zero partition
\begin{align}
     \Rd &= \JR_1  \dot{q}_1 + \cdots +   \JR_n  \dot{q}_n \label{eq:Rdot}\\
     \td &= \Jt_1  \dot{q}_1 + \cdots +   \Jt_n  \dot{q}_n \label{eq:tdot}
\end{align}
where each term represents the contribution to end-effector velocity due to the motion of the corresponding joint.

Taking (\ref{eq:tdot}) first, we can simply write
\begin{align}
     \td &= 
     \begin{pmatrix}
          \Jt_1 & \cdots  & \Jt_n 
     \end{pmatrix} 
     \begin{pmatrix}
          \dot{q}_1 \\ \vdots \\  \dot{q}_n 
     \end{pmatrix} \nonumber \\
     &= 
     \Jv(\q) \qd  \label{eq:td}
\end{align}
where $\Jv(\q) \in \mathbb{R}^{3 \times n}$ is the translational part of the manipulator Jacobian matrix.

Rotation rate is slightly more complex, but using the identity $\Rd = \skx{\w} \R$ where $\vec{\omega} \in \mathbb{R}^{3}$ is the angular velocity, and $\skx{\w} \in \bf{so}(3)$ is a skew-symmetric matrix, we can rewrite (\ref{eq:Rdot}) as
\begin{equation} \label{eq:skxr}
     \skx{\w} \R = \JR_1  \dot{q}_1 + \cdots +   \JR_n  \dot{q}_n 
\end{equation}
and rearrange to 
\begin{equation*}
     \skx{\w}  = (\JR_1 \R^\top) \dot{q}_1 + \cdots +   (\JR_n \R^\top) \dot{q}_n  \in \bf{so}(3).
\end{equation*}
This $3\times 3$ matrix equation therefore has only three unique equations so applying the inverse skew operator to both sides we have
\begin{align} \label{eq:Rd}
     \vec{\omega}  
     &= 
     \iskx{\JR_1 \R^\top } \dot{q}_1 + \cdots + 
     \iskx{\JR_n \R^\top} \dot{q}_n \nonumber \\
     &= 
     \bigg(
     \begin{matrix}
          \iskx{\JR_1 \R^\top} &  \cdots  &  \iskx{\JR_n \R^\top}
     \end{matrix}
     \bigg)
     \begin{pmatrix}
          \dot{q}_1 \\ 
          \vdots \\  
          \dot{q}_n 
     \end{pmatrix} \nonumber \\
     &= \Jw(\q) \qd 
\end{align}
where $\Jw(\q)  \in \mathbb{R}^{3 \times n}$ is the rotational part of the manipulator Jacobian.

Combining \eq{eq:td} and \eq{eq:Rd} we can write
\begin{equation} \label{eq:jacob}
    {^0\vec{\nu}} =
     \begin{pmatrix} 
          \vec{v} \\ 
          \w 
     \end{pmatrix} = 
     \begin{pmatrix} 
          \Jv(\q) \\ 
          \Jw(\q) 
     \end{pmatrix} \qd  
\end{equation}
which expresses end-effector spatial velocity ${^0\vec{\nu}} = (v_x, \ v_y, \ v_z, \ \omega_x, \ \omega_y, \ \omega_z)$ in the world frame in terms of joint velocity and
\begin{equation} \label{eq:JT}
     {^0\J}(\q) = 
     \begin{pmatrix} 
          \Jv(\q) \\ 
          \Jw(\q) 
     \end{pmatrix} \in \mathbb{R}^{6\times n}
\end{equation}
is the manipulator Jacobian matrix expressed in the world-coordinate frame. The Jacobian expressed in the end-effector frame is
\begin{equation}
    {^e\J}(\q) =
    \begin{pmatrix} 
        {{^0\mat{R}}_e^\top} & \mathbf{0} \\ 
        \mathbf{0} & {{^0\mat{R}}_e^\top}
     \end{pmatrix}
     {^0\J}(\q)
\end{equation}
where ${^0\mat{R}}_e$ is a rotation matrix representing the orientation of the end-effector in the world frame.

Combining (\ref{eq:jacob}) and (\ref{eq:JT}) we write
\begin{equation} \label{eq:j0}
    {^0\vec{\nu}}
     = {^0\J}(\q) \, \qd
\end{equation}
which provides the derivative of the left side of (\ref{m:ets1}).
However, in order to compute (\ref{eq:j0}), we need to first compute (\ref{eq:partition}), that is $\frac{\partial \mat{T}}{\partial q_j}$.


\subsection*{First Derivative of an Elementary\\Transform} \label{sec:dE}

Before differentiating the ETS to find the manipulator Jacobian, it is useful to consider the derivative of a single Elementary Transform.

\paragraph{Derivative of a Pure Rotation}

The derivative of a rotation matrix with respect to the rotation angle $\theta$ is required when considering a revolute joint and
can be shown to be
\begin{align} \label{eq:dr1}
    \frac{\mathrm{d}\mat{R}(\theta)}
         {\mathrm{d} \theta} 
    &= \skx{\evec{\omega}} \mat{R} \big( \theta(t) \big)
\end{align}
where the unit vector $\evec{\omega}$ is the joint rotation axis.

The rotation axis $\evec{\omega}$ can be recovered using the inverse skew operator
\begin{align}
    \evec{\omega} &= \iskx{ 
        \frac{\mathrm{d} \matfn{R}{\theta}}
             {\mathrm{d} \theta}
    \mat{R}  \big( \theta(t)\big)^\top } \label{m:dw1} 
\end{align}
since $ \mat{R} \in \SO{3}$, then $\mat{R}^{-1} = \mat{R}^\top$.

For an ETS, we only need to consider the elementary rotations $\mat{R}_x$, $\mat{R}_y$, and $\mat{R}_z$ as shown in Figure \ref{fig:et} which are embedded within \SE{3}, as $\TR{x}$, $\TR{y}$, and $\TR{z}$ i.e. pure rotations with no translational component. We can show that the derivative of each elementary rotation with respect to a rotation angle is
\begin{align}
    \dfrac{\mathrm{d} \TR{x}(\theta)}
          {\mathrm{d} \theta}
    &=     
    \begin{pmatrix}
        0 & 0 & 0 & 0 \\
        0 & 0 & -1 & 0 \\
        0 & 1 & 0 & 0 \\
        0 & 0 & 0 & 0 
    \end{pmatrix} \TR{x}(\theta) = \big[ \evec{R}_x \big] \TR{x}(\theta), \label{eq:ERx} \\
    \dfrac{\mathrm{d} \TR{y}(\theta)}
          {\mathrm{d} \theta}
    &=     
    \begin{pmatrix}
        0 & 0 & 1 & 0 \\
        0 & 0 & 0 & 0 \\
        -1 & 0 & 0 & 0 \\
        0 & 0 & 0 & 0 
    \end{pmatrix} \TR{y}(\theta) = \big[ \evec{R}_y \big] \TR{y}(\theta), \label{eq:ERy} \\
    \dfrac{\mathrm{d} \TR{z}(\theta)}
          {\mathrm{d} \theta}
    &=     
    \begin{pmatrix}
        0 & -1 & 0 & 0 \\
        1 & 0 & 0 & 0 \\
        0 & 0 & 0 & 0 \\
        0 & 0 & 0 & 0  
    \end{pmatrix} \TR{z}(\theta) = \big[ \evec{R}_z \big] \TR{z}(\theta). \label{eq:ERz}
\end{align}

If a joint's defined positive rotation is a negative rotation about the axis, 
as is $\mat{E}_7$ and $\mat{E}_{11}$ in the ETS of the Panda shown in Figure \ref{fig:cover}, then
$\big[ \evec{R}_i \big]^\top  \mat{T}_{\mat{R}_i}(\theta)^\top$
is used to calculate the derivative.

\paragraph{Derivative of a Pure Translation}

Consider the three elementary translations $\Tt{}$ shown in Figure \ref{fig:et}.

The derivative of a homogeneous transformation matrix with respect to translation is required when considering a prismatic joint. For an ETS, these translations are embedded in \SE{3} as $\Tt{x}$, $\Tt{y}$, and $\Tt{z}$ which are pure translations with zero rotational component. We can show that the derivative of each elementary translation with respect to a translation is 
\begin{align}
    \dfrac{\mathrm{d} \Tt{x}(d)}
          {\mathrm{d} d}
    &=     
    \begin{pmatrix}
        0 & 0 & 0 & 1 \\
        0 & 0 & 0 & 0 \\
        0 & 0 & 0 & 0 \\
        0 & 0 & 0 & 0 
    \end{pmatrix} = \sk{\evec{t}_x}, \label{eq:ETx}\\
    \dfrac{\mathrm{d} \Tt{y}(d)}
          {\mathrm{d} d}
    &=     
    \begin{pmatrix}
        0 & 0 & 0 & 0 \\
        0 & 0 & 0 & 1 \\
        0 & 0 & 0 & 0 \\
        0 & 0 & 0 & 0 
    \end{pmatrix} = \sk{\evec{t}_y}, \label{eq:ETy}\\
    \dfrac{\mathrm{d} \Tt{z}(d)}
          {\mathrm{d} d}
    &=     
    \begin{pmatrix}
        0 & 0 & 0 & 0 \\
        0 & 0 & 0 & 0 \\
        0 & 0 & 0 & 1 \\
        0 & 0 & 0 & 0 
    \end{pmatrix} = \sk{\evec{t}_z}. \label{eq:ETz}
\end{align}

No changes are required if a joint's defined positive translation is a negative translation along the axis.

\begin{figure}[!t]
    \centering
    \includegraphics[height=2.4cm]{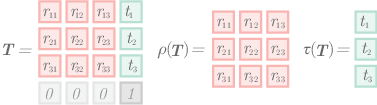}
    \caption{
        Visualisation of a homogeneous transformation matrix (the derivatives share the form of $\mat{T}$, except will have a 0 instead of a 1 located at $\mat{T}_{44}$). Where the matrix $\rho(\mat{T}) \in \mathbb{R}^{3 \times 3}$ of red boxes forms the rotation component, and the vector $\tau(\mat{T}) \in \mathbb{R}^{3}$ of blue boxes form the translation component. The rotation component can be extracted through the function $\rho(\cdot) : \mathbb{R}^{4\times 4} \mapsto  \mathbb{R}^{3\times 3}$, while the translation component can be extracted through the function $\tau(\cdot) : \mathbb{R}^{4\times 4} \mapsto  \mathbb{R}^{3}$.
    }
    \label{fig:transform}
\end{figure}


\subsection*{The Manipulator Jacobian} \label{sec:dETS}

Now we can calculate the derivative of an ETS. To find out how the $j^{th}$ joint affects the end-effector pose, apply the chain rule to (\ref{m:ets1})
\begin{align} \label{eq:pdr}
    \frac{\partial \matfn{T}{\vec{q}}}
         {\partial q_j} 
    &=
    \frac{\partial} 
         {\partial q_j} \nonumber 
    \left(
        \mat{E}_1(\eta_1) \mat{E}_2(\eta_2) \hdots \mat{E}_M(\eta_M) 
    \right)\\
    &= 
    \prod_{i=1}^{\mu(j)-1} \!\!\!\mat{E}_i(\eta_i) 
    \ \
    \frac{\mathrm{d} \mat{E}_{\mu(j)}(q_j)} 
         {\mathrm{d} q_j}
    \prod_{i=\mu(j)+1}^{M} \!\!\!\!\!\!\mat{E}_i(\eta_i).
\end{align}

The derivative of the elementary transform with respect to a joint coordinate in (\ref{eq:pdr}) is obtained using one of (\ref{eq:ERx}), (\ref{eq:ERy}), or (\ref{eq:ERz}) for a revolute joint, or one of (\ref{eq:ETx}), (\ref{eq:ETy}), or (\ref{eq:ETz}) for a prismatic joint.

Using (\ref{eq:Rd}) with (\ref{eq:pdr}) we can form the angular velocity component of the $j^{th}$ column of the manipulator Jacobian
\begin{align} \label{eq:jwj1}
    \Jwx{j}(\q) 
    &=
    \iskx{
        \rho
        \left(
            \frac{\partial \matfn{T}{\vec{q}}}
                {\partial q_j} 
        \right)
        \rho
        \left(
            \matfn{T}{\vec{q}}
        \right)^\top
    }
\end{align}
where the $\rho(\cdot)$ operation is defined in Figure \ref{fig:transform}. Using (\ref{eq:td}) with (\ref{eq:pdr}), the translational velocity component of the $j^{th}$ column of the manipulator Jacobian is
\begin{align} \label{eq:jvj1}
    \Jvx{j}(\q) 
    &=
    \tau
    \left(
        \frac{\partial \matfn{T}{\vec{q}}}
            {\partial q_j} 
    \right)
\end{align}
where the $\tau(\cdot)$ operation is defined in Figure \ref{fig:transform}.

\begin{figure}[!t]
    \centering
    \includegraphics[height=5cm]{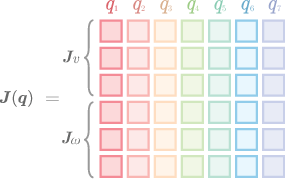}
    \caption{
        Visualisation of the Jacobian $\J(\q)$ of a 7-jointed manipulator. Each column describes how the end-effector pose changes
        due to motion of the corresponding joint.
        The top three rows $\Jv$  correspond to the linear velocity of the end-effector while the bottom three rows $\Jw$ correspond to the angular velocity of the end-effector.
    }
    \label{fig:jacobian}
\end{figure}

Stacking the translational and angular velocity components, the $j^{th}$ column of the manipulator Jacobian becomes 
\begin{equation}
     \J_j(\q) = 
     \begin{pmatrix} 
          \Jvx{j}(\q) \\ 
          \Jwx{j}(\q) 
     \end{pmatrix} \in \mathbb{R}^{6}
\end{equation}
where the full manipulator Jacobian is 
\begin{equation}
    \J(\q) = 
    \begin{pmatrix} 
        \J_1(\q) & \cdots & \J_n(\q)
    \end{pmatrix} \in \mathbb{R}^{6 \times n}
\end{equation}
and we display a visualisation of the general structure of the manipulator Jacobian in Figure \ref{fig:jacobian}.

\subsection*{Fast Manipulator Jacobian}

Calculating the manipulator Jacobian using (\ref{eq:jwj1}) and (\ref{eq:jvj1}) is easy to understand, but has $\mathcal{O}(n^2)$ time complexity -- we can do better.

Expanding (\ref{eq:jwj1}) using (\ref{eq:pdr}) and simplifying using $\R \R^\top = \mat{1}$ gives
\begin{align} \label{eq:jwj2}
    \Jwx{j}(\q) 
    &=
    \vee_\times \bigg(
        \rho
        \left(
            ^0\T_j
        \right)
        \rho
        \left(
            \sk{\evec{G}_{\mu(j)}}
        \right)
        \left(
            \rho(^0\T_j)^\top
        \right)
    \bigg)
\end{align}
where $^0\mat{T}_j$ represents the transform from the base frame to frame $\{j\}$ as described by (\ref{m:joint_pose}), and $[\evec{G}_{\mu(j)}]$ corresponds to one of the six elementary transform derivatives from equations (\ref{eq:ERx})-(\ref{eq:ERz}) and (\ref{eq:ETx})-(\ref{eq:ETz}). 

\begin{figure}[!t]
    \centering
    \includegraphics[height=5cm]{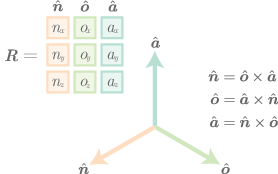}
    \caption{
        The two vector representation of a rotation matrix $\R \in \SO{3}$. The rotation matrix $\R$ describes the coordinate frame in terms of three orthogonal vectors $\evec{n}$, $\evec{o}$, and $\evec{a}$ which are the axes of the rotated frame expressed in the reference coordinate frame $\evec{x}$, $\evec{y}$, and $\evec{z}$. As shown above, each of the vectors $\evec{n}$, $\evec{o}$, and $\evec{a}$ can be calculated using the cross product of the other two.
    }
    \vspace{-10pt}
    \label{fig:rotation}
\end{figure}

In the case of a prismatic joint, $\rho(\evec{G}_{\mu(j)})$ will be a $3 \times 3$ matrix of zeros which results in zero angular velocity. In the case of a revolute joint, the angular velocity is parallel to the axis of joint rotation.
\begin{align}
    \Jwx{j}(\q) 
    &=
    \left\{ 
    \begin{matrix*}[l]
        \evec{n}_j & \mbox{if} \ \ \mat{E}_m = \TR{x}\\
        \evec{o}_j & \mbox{if} \ \ \mat{E}_m = \TR{y}\\
        \evec{a}_j & \mbox{if} \ \ \mat{E}_m = \TR{z}\\
    \begin{pmatrix} 0 & 0 & 0 \end{pmatrix}^\top & \mbox{if} \ \ \mat{E}_m = \Tt{}\\
    \end{matrix*}
    \right.
\end{align}
where $\begin{pmatrix} \evec{n}_j & \evec{o}_j & \evec{a}_j \end{pmatrix} = \rho(^0\T_j)$ are the columns of a rotation matrix as shown in Figure \ref{fig:rotation}.

Expanding (\ref{eq:jvj1}) using (\ref{eq:pdr}) provides
\begin{align} \label{eq:jwj22}
    \Jvx{j}(\q) 
    &=
        \tau
        \left(
            ^0\T_j
            \sk{\evec{G}_{\mu(j)}}
            {^j\T}_{e}
        \right)
\end{align}
which reduces to 
\begin{align}
    \Jvx{j}(\q) 
    &=
    \left\{ 
    \begin{matrix*}[l]
        \evec{a}_j y_e - \evec{o}_j z_e & \mbox{if} \ \ \mat{E}_m = \TR{x}\\
        \evec{n}_j z_e - \evec{a}_j x_e & \mbox{if} \ \ \mat{E}_m = \TR{y}\\
        \evec{o}_j x_e - \evec{y}_j y_e & \mbox{if} \ \ \mat{E}_m = \TR{z}\\
        \evec{n}_j                      & \mbox{if} \ \ \mat{E}_m = \Tt{x}\\
        \evec{o}_j                      & \mbox{if} \ \ \mat{E}_m = \Tt{y}\\
        \evec{a}_j                      & \mbox{if} \ \ \mat{E}_m = \Tt{z}\\
    \end{matrix*}
    \right.
\end{align}
where $\begin{pmatrix}x_e & y_e & z_e \end{pmatrix}^\top = \tau({^j\T}_{e})$.

This simplification reduces the time complexity of computation of the manipulator Jacobian to $\mathcal{O}(n)$.


\section*{Manipulator Jacobian\\Applications}

The manipulator Jacobian is widely used in robotic control algorithms and the remainder of this article details several applications of the manipulator Jacobian.

\subsection*{Resolved-Rate Motion Control}

Resolved-rate motion control (RRMC), a direct application of the first-order differential equation we generated in (\ref{eq:j0}), is a simple and elegant method to generate joint velocities which create a desired Cartesian end effector motion \cite{rrmc}. The ability to instantaneously change the end-effector velocity is at the heart of sensor-based reactive control. By ``closing the loop'' on sensory information a robot can adapt to changes in its environment as new knowledge is acquired by sensors. A well-known example is visual servoing, a technique that pairs robotic vision with RRMC to guide the manipulator to an image-based goal \cite{vs}. RRMC is also foundational for numerical inverse kinematics which we will visit in the next section \cite{ik1, ik3}. Of course, this introductory technique, while still highly relevant and powerful, can be extended and enhanced to improve robustness and consider constraints such as joint limits and environmental collisions \cite{mmc, mmc2, neo, re0, pot, mansard2009versatile}.

We first re-arrange (\ref{eq:j0}) 
\begin{equation}  \label{eq:rrmc1}
    \dvec{q} = {^0\mat{J}}(\vec{q})^{-1} \ {^0\vec{\nu}}
\end{equation}
which can only be solved when $\matfn{J}{\vec{q}}$ is square (and non-singular), which is when the robot has six degrees-of-freedom.

For redundant robots, there is no unique solution for (\ref{eq:rrmc1}). Consequently, the most common approach is to use the Moore-Penrose pseudoinverse
\begin{equation} \label{eq:rrmc2}
    \dvec{q} = {^0\mat{J}}(\vec{q})^{+} \ {^0\vec{\nu}}.
\end{equation}

The pseudoinverse will return joint velocities with the minimum velocity norm of the possible solutions, although note an important caveat on mixed units described in Excurse \ref{fig:note1}.

Immediately from this, we can construct a primitive open-loop velocity controller. At each time step, we must calculate the manipulator Jacobian $\matfn{J}{\vec{q}}$ which is a function of the robot's current configuration $\vec{q}$. Then we set ${^0\vec{\nu}}$ to the desired spatial velocity of the end-effector.

\begin{figure}[!t]
    \centering
    \includegraphics[height=4.95cm]{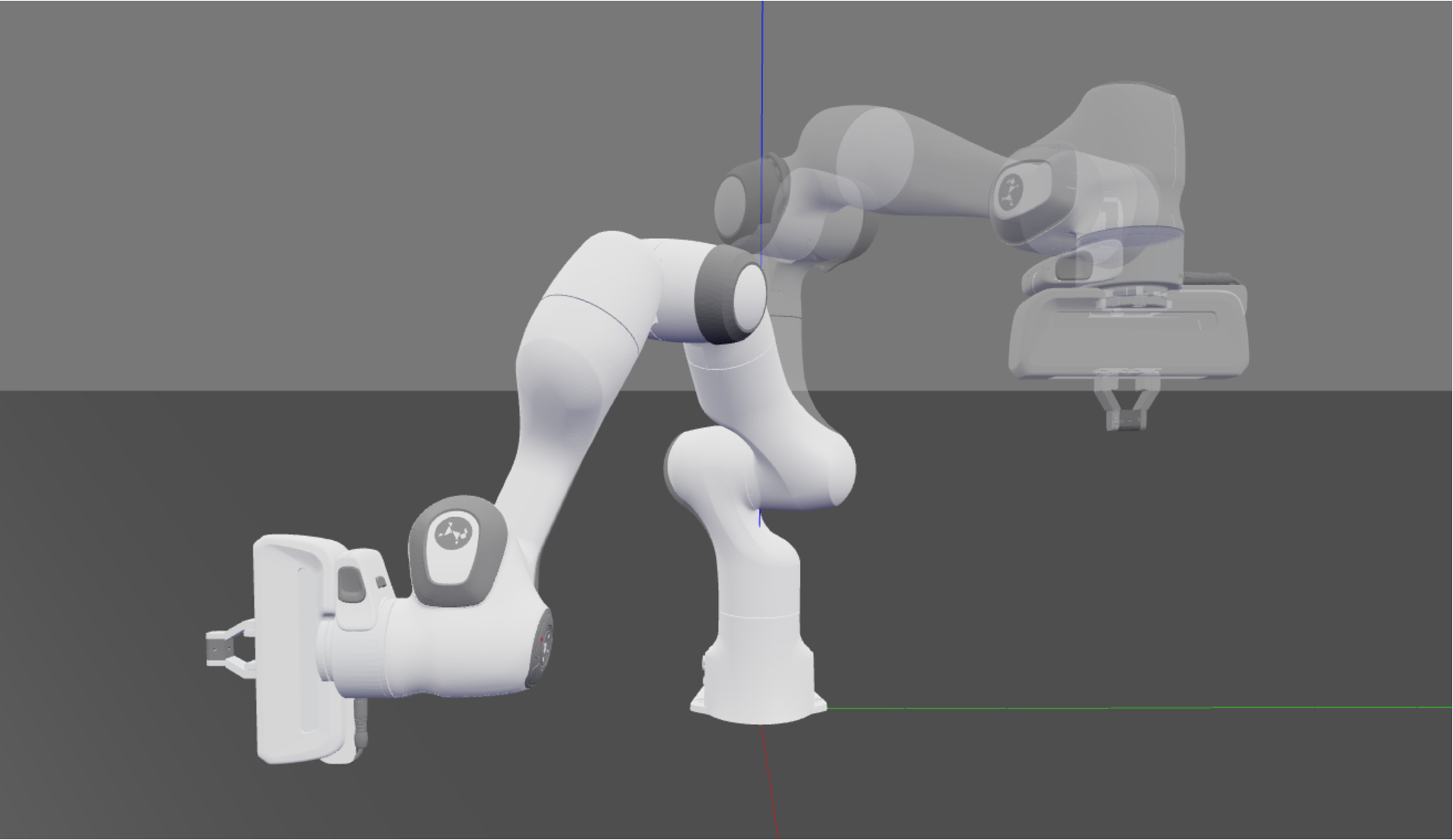}
    \caption{
        Visualisation of a Panda robot which has been controlled by a position-based servoing control scheme utilising resolved-rate motion control. The end-effector's pose has both been translated and rotated to reach the desired pose.
    }
    \label{fig:rrmc1}
\end{figure}

A more useful application of RRMC is to employ it in a closed-loop pose controller which we denote as position-based servoing (PBS). Using this method we can make the end-effector travel in a straight line, in the robot's task space, towards some desired end-effector pose as displayed in Figure \ref{fig:rrmc1}. The PBS scheme relies on an error vector which represents the translation and rotation from the end-effector's current pose to the desired pose
\begin{equation} \label{eq:pbs1}
    \vec{e} =
    \begin{pmatrix}
        \tau\left(\mat[0]{T}_{e^*}\right) - \tau\left(\mat[0]{T}_{e}\right) \\
        \alpha
        \left(
            \rho\left(\mat[0]{T}_{e^*}\right)
            \rho\left(\mat[0]{T}_{e}\right)^\top
        \right)
    \end{pmatrix} \in \mathbb{R}^6
\end{equation}
where the top three rows correspond to the translational error in the world frame, the bottom three rows correspond to the rotational error in the world frame,
${^0\T_e} = {\cal K}(\q)$
is the forward kinematics of the robot which represents the end-effector pose in the base frame of the robot, $\mat[0]{T}_{e^*}$ is the desired end-effector pose in the base frame of the robot ($\cdot^*$ denotes desired), and $\alpha(\cdot) : \SO{3} \mapsto \mathbb{R}^3$ transforms a rotation matrix to its Euler vector equivalent \cite{ik3}. The Euler vector is a form of angle-axis representation and specifies an axis of rotation and the angle of rotation about that axis.
\begin{excurse}[!t]

    \noindent\textcolor{lineblue}{\rule{\linewidth}{2pt}}
    \vspace{-13pt}
    
    \noindent\colorbox{boxblue}{
        
        \parbox{\linewidth-12.5pt}{
            \vspace{4pt}
            \caption{
                \usefont{OT1}{ugq}{n}{n}{Inhomogeneous Units}
            }
            \label{fig:note1}
            \vspace{-5pt}

            \color{dgrey}
            Vector norms can be problematic when they contain mixed units.  For example  $\| \vec{e} \|$ and $\| \posedot \|$ are non-homogeneous as they both combine translational and rotational units. For manipulators with mixed prismatic and revolute joints, so too does $\| \dvec{q} \|$.  Solutions include scaling \cite{stocco1999use} to ensure that the translational and rotational values have commensurate magnitude, or to consider translational and rotational components separately.
    
            \vspace{3pt}
        }
    }
    
    \vspace{-1pt}
    \noindent\textcolor{lineblue}{\rule{\linewidth}{2pt}} 
\end{excurse}

If $\mat{R}$ is not a diagonal matrix then the Euler vector equivalent of $\mat{R}$ is calculated as
\begin{align}
    \alpha
    \left(
        \mat{R}
    \right) &=
    \dfrac
    {\mbox{atan2}
    \left(
        \lVert\vec{l}\rVert, r_{11} + r_{22} + r_{33} - 1
    \right)}
    {\lVert\vec{l}\rVert}\vec{l}.
\end{align}
where using the convention from Figure \ref{fig:transform} we define $\rho(\mat{T}) = \mat{R} = \{r_{ij}\}$ and
\begin{align}
    \vec{l} =
    \begin{pmatrix}
        r_{32} - r_{23} \\
        r_{13} - r_{31} \\
        r_{21} - r_{12} \\
    \end{pmatrix}.
\end{align}

If $\mat{R}$ is a diagonal matrix then we use different formulas. For the case where $\begin{pmatrix}r_{11} & r_{22} & r_{33}\end{pmatrix} = \begin{pmatrix}1&1&1\end{pmatrix}$ then $\alpha(\mat{R}) = \begin{pmatrix}0 & 0 & 0\end{pmatrix}^\top$ otherwise
\begin{align}
    \alpha
    \left(
        \mat{R}
    \right) &=
    \frac{\pi}{2}
    \begin{pmatrix}
        r_{11} + 1 \\
        r_{22} + 1 \\
        r_{33} + 1 \\
    \end{pmatrix}.
\end{align}

To construct the PBS scheme we take the error term from (\ref{eq:pbs1}) to set $\vec{\nu}$ in (\ref{eq:rrmc1}) (or (\ref{eq:rrmc2}) for robots with 7+ degrees of freedom) at each time step
\begin{align} \label{eq:pbs2}
    \vec{\nu} = \mat{k} \vec{e}
\end{align}
where $\mat{k}$ is a proportional gain with units $s^{-1}$ that controls the rate of convergence to the goal
and is typically a diagonal matrix to set gains for each task-space DoF
\begin{align}
    \mat{k} =
    \mbox{diag}
    (
        k_t, \ k_t, \ k_t, \ k_r, \ k_r, \ k_r
    )
\end{align}
where $k_t$ is the translational motion gain and $k_r$ is the rotational motion gain. The inverse of the gain, $\mat{k}^{-1}$, corresponds to the time constant of the scheme. Note that since the error $\vec{e}$ is in the base frame of the robot we must use the base-frame manipulator Jacobian $^0\mat{J}(\vec{q})$ in (\ref{eq:rrmc1}) or (\ref{eq:rrmc2}).

The control scheme we have just described will cause the error to asymptotically decrease to zero. 
For real applications, this is slow and impractical.
We can improve this by increasing $k_t$ and $k_r$ and capping the end-effector velocity norm
${\lVert \vec{\nu} \rVert}$
at some value $\nu_{m}$, before stopping when the error norm ${\lVert \vec{e} \rVert}$ drops below some value $e_m$
\begin{align} \label{eq:rrmc3}
    \vec{\nu}
    &=
    \left\{ 
    \begin{matrix*}[l]
        \mat{k}\vec{e} \dfrac{\nu_{m}}{\lVert \mat{k}\vec{e} \rVert} & \mbox{if} \ \ \lVert \mat{k}\vec{e} \rVert > \nu_{m} \\
        \ \mat{k}\vec{e} & \mbox{if} \ \ \nu_{m} \geq \lVert \mat{k}\vec{e} \rVert > e_m \\
        \vec{0} & \mbox{otherwise.}
    \end{matrix*}
    \right.
\end{align}

\begin{figure}[!t]
    \centering
    \includegraphics[height=5.1cm]{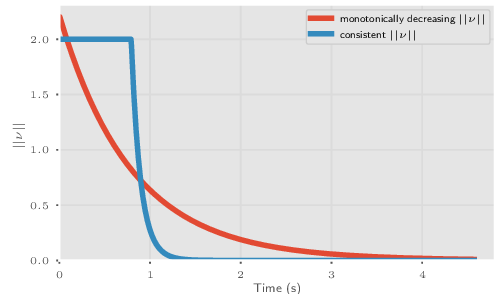}
    \caption{
        Comparison of the Euclidean distance error from the end-effectors pose to the desired pose using the monotonically decreasing $\lVert \vec{\nu} \rVert$ method from (\ref{eq:pbs2}) and the consistent $\lVert \vec{\nu} \rVert$ method from (\ref{eq:rrmc3}).
    }
    \label{fig:rrmc2}
\end{figure}

This will cause ${\lVert \vec{\nu} \rVert}$ to be consistent 
until $e_m$ is reached and subsequently
${\lVert \vec{\nu} \rVert}$ asymptotically decreases to safely stop the robot. The effect of this is displayed in Figure \ref{fig:rrmc2} where $\nu_m$ has been set to 2.0.
This approach involves combining mixed units, see Excurse \ref{fig:note1}.

This particular approach may cause unintuitive results as both 
${\lVert \vec{e} \rVert}$ and ${\lVert \vec{\nu} \rVert}$
are non-homogeneous -- they combine translational and rotational units.
Therefore, a more intuitive approach would be to modify (\ref{eq:rrmc3}) to adjust the translational and angular velocity profiles separately.
Alternate velocity profiles, such as a linearly decreasing velocity norm, are possible through modification of (\ref{eq:rrmc3}).

Although we have the safety mechanism of capping ${\lVert \vec{\nu} \rVert}$ at $\nu_m$, care must still be taken to not increase $k_t$ and $k_r$ without due consideration. Well-known problems caused by an excessive gain value, such as overshoot and system instability, may become apparent, especially on a real-world robot (which has non-modelled dynamics including flexibility, communication delay, and non-linearities).

For a \emph{reactive} task, where a dynamic environment is causing the goal pose to change, simply update the desired end-effector pose $\mat[0]{T}_{e^*}$ in (\ref{eq:pbs1}) at each time step, based on sensory information. A technique such as position-based visual servoing could be used to achieve this \cite{vs}. Refer to \href{https://github.com/jhavl/dkt}{Notebook 3} where we use the Swift simulator to show this in action.


\subsection*{Numerical Inverse Kinematics}

Inverse kinematics is the problem of determining the corresponding joint coordinates, given some end-effector pose.
There are two approaches to solving inverse kinematics: analytical and numerical.

Analytical formulas must be pre-generated for a given manipulator and in some cases may not exist. The  IKFast program provided as a part of OpenRAVE, pre-compiles analytic solutions for a given manipulator in optimised C++ code \cite{openrave}. After this initial step, inverse kinematics can be computed rapidly, taking as little as 5 microseconds. However, analytic solutions generally can not optimise for additional criteria such as joint limits.

Numerical inverse kinematics use an iterative technique and can additionally consider extra constraints such as collision avoidance, joint limit avoidance, or manipulability \cite{ik1, ik3, ik4}. In this section, we first construct a primitive numerical inverse kinematics solver, before showing how it can be improved using advanced optimisation techniques.

\newcolumntype{M}[1]{>{\centering\arraybackslash}m{#1}}
\begin{table*}[b!]
    \centering
    \small
    \renewcommand{\arraystretch}{1.3}

    \caption{Numerical IK Methods Compared over $10\,000$ Problems on a 6 DoF UR5 Manipulator}
    \label{tab:ik}

    \begin{tabular}{ m{3.8cm} | M{1.3cm} | M{1.2cm} | M{0.9cm} | M{1.0cm}  | M{1.3cm} | M{1.3cm} | M{1.4cm} | M{1.4cm}}
    \hline
    \centering{Method} & Searches Allowed & Iter. Allowed & Mean Iter. & Median Iter. & Infeasible Count & Infeasible \% & Mean Searches & Max Searches\\
    \hline\hline
                              NR &     1   &   500   &  21.34   &    16.0   &   1093   &   10.93\%   &     1.0    &    1.0 \\
                              GN &     1   &   500   &   21.6   &    16.0   &   1078   &   10.78\%   &     1.0    &    1.0 \\
                NR Pseudoinverse &     1   &   500   &  21.24   &    16.0   &   1100   &    11.0\%   &     1.0    &    1.0 \\
                GN Pseudoinverse &     1   &   500   &  21.72   &    16.0   &   1090   &    10.9\%   &     1.0    &    1.0 \\
    LM (Wampler $\lambda=1e- 4$) &     1   &   500   &   20.1   &    14.0   &    934   &    9.34\%   &     1.0    &    1.0 \\
    LM (Wampler $\lambda=1e- 6$) &     1   &   500   &  29.84   &    17.0   &    529   &    5.29\%   &     1.0    &    1.0 \\
         LM (Chan $\lambda$=1.0) &     1   &   500   &  16.58   &    14.0   &   1011   &   10.11\%   &     1.0    &    1.0 \\
         LM (Chan $\lambda$=0.1) &     1   &   500   &   9.43   &    9.0    &   963    &    9.63\%   &     1.0    &    1.0 \\
        LM (Sugihara $w_n=1e-3$) &     1   &   500   &  20.54   &    15.0   &   1024   &   10.24\%   &     1.0    &    1.0 \\
        LM (Sugihara $w_n=1e-4$) &     1   &   500   &  17.01   &    14.0   &   1011   &   10.11\%   &     1.0    &    1.0 \\
    \hline
                              NR &   100   &    30   &  30.16   &    18.0   &      0   &     0.0\%   &    1.47    &   25.0 \\
                              GN &   100   &    30   &  30.33   &    18.0   &      0   &     0.0\%   &    1.48    &   23.0 \\
                NR Pseudoinverse &   100   &    30   &  30.27   &    18.0   &      0   &     0.0\%   &    1.47    &   20.0 \\
                GN Pseudoinverse &   100   &    30   &  30.65   &    18.0   &      0   &     0.0\%   &    1.49    &   20.0 \\
    LM (Wampler $\lambda=1e- 4$) &   100   &    30   &  25.23   &    15.0   &      0   &     0.0\%   &    1.35    &   17.0 \\
    LM (Wampler $\lambda=1e- 6$) &   100   &    30   &  29.3    &    18.0   &      0   &     0.0\%   &    1.45    &   24.0 \\
         LM (Chan $\lambda$=1.0) &   100   &    30   &  22.6    &    15.0   &      0   &     0.0\%   &    1.25    &   18.0 \\
         LM (Chan $\lambda$=0.1) &   100   &    30   &  15.33   &    9.0    &      0   &     0.0\%   &     1.2    &   18.0 \\
        LM (Sugihara $w_n=1e-3$) &   100   &    30   &  26.49   &    16.0   &      0   &     0.0\%   &    1.35    &   18.0 \\
        LM (Sugihara $w_n=1e-4$) &   100   &    30   &  23.04   &    15.0   &      0   &     0.0\%   &    1.26    &   18.0 \\
    \hline
    \end{tabular}
\end{table*}

The Newton-Raphson (NR) method for inverse kinematics is remarkably similar to RRMC in a position-based servoing scheme. However, instead of sending joint velocities to the manipulator, the \emph{joint velocities} update the configuration of a virtual manipulator until the goal pose is reached. To find the joint coordinates which correspond to some end-effector pose $^0\mat{T}_{e^*}$, the NR method seeks to minimise an error function
\begin{align} \label{eq:eqp}
    E = \frac{1}{2} \vec{e}^{\top} \mat{W}_e \vec{e}
\end{align}
where $\vec{e}$ is defined in (\ref{eq:pbs1}), and $\mat{W}_e = \mbox{diag}(\vec{w_e})\big(\vec{w_e} \in (\mathbb{R}^+)^n \big)$ is a diagonal weighting matrix which prioritises the corresponding error term. To achieve this, we iterate upon the following
\begin{align} \label{eq:ik1}
    \vec{q}_{k+1} = \vec{q}_k + {^0\mat{J}(\vec{q}_k)}^{-1} \vec{e}_k.
\end{align}

With this approach, $^0\mat{J}(\vec{q})$ must be square (and non-singular) to be invertible and is therefore limited to manipulators with 6 joints.

When using the NR method, the initial joint coordinates $q_0$ should correspond to a non-singular manipulator pose, since it uses the manipulator Jacobian.
When the problem is solvable, it converges very quickly.
However, this method frequently fails to converge on the goal. 
We can improve the solvability of the problem and add compatibility with manipulators with greater than six joints, by using the Gauss-Newton (GN) method
\begin{align} \label{eq:ik2}
    \vec{q}_{k+1} &= \vec{q}_k +
    \left(
    {\mat{J}(\vec{q}_k)}^\top
    \mat{W}_e \
    {\mat{J}(\vec{q}_k)}
    \right)^{-1}
    \vec{g}_k \\
    \vec{g}_k &=
    {\mat{J}(\vec{q}_k)}^\top
    \mat{W}_e
    \vec{e}_k
\end{align}
where $\mat{J} = {^0\mat{J}}$ is the base-frame manipulator Jacobian. If $\mat{J}(\vec{q}_k)$ is non-singular, and $\mat{W}_e = \mat{1}_n$, then (\ref{eq:ik2}) provides the pseudoinverse solution to (\ref{eq:ik1}). However, if $\mat{J}(\vec{q}_k)$ is singular, (\ref{eq:ik2}) cannot be computed and the GN solution is infeasible.

Most linear algebra libraries (including the Python numpy library) implement the pseudoinverse using singular value decomposition, which is robust to singular matrices. Therefore, we can make both solvers more robust by using the pseudoinverse instead of the normal inverse in (\ref{eq:ik1}) and (\ref{eq:ik2}).

However, the computation is still unstable near singular points. We can further improve the solvability through the Levenberg-Marquardt (LM) method
\begin{align} \label{eg:ik4}
    \vec{q}_{k+1} 
    &= 
    \vec{q}_k +
    \left(
        \mat{A}_k
    \right)^{-1}
    \vec{g}_k \\
    \mat{A}_k
    &=
    {\mat{J}(\vec{q}_k)}^\top
    \mat{W}_e \
    {\mat{J}(\vec{q}_k)}
    +
    \mat{W}_n
\end{align}
where $\mat{W}_n = \mbox{diag}(\vec{w_n})\big(\vec{w_n} \in (\mathbb{R}^+)^n \big)$ is a diagonal damping matrix. The damping matrix ensures that $\mat{A}_k$ is non-singular and positive definite. The performance of the LM method largely depends on the choice of $\mat{W}_n$. Wampler \cite{wampler} proposed $\vec{w_n}$ to be a constant, Chan and Lawrence \cite{chan} proposed a damped least-squares method with
\begin{align}
    \mat{W}_n
    &=
    \lambda E_k \mat{1}_n
\end{align}
where $\lambda$ is a constant which does not have much influence on performance, and $E_k$ is the error value from (\ref{eq:eqp}) at time step $k$. Sugihara \cite{ik3} proposed 
\begin{align}
    \mat{W}_n
    &=
    E_k \mat{1}_n + \mbox{diag}(\hvec{w}_n)
\end{align}
where $\hvec{w}_n \in \mathbb{R}^n$, $\hat{w}_{n_i} = l^2 \sim 0.01 l^2$, and $l$ is the length of a typical link within the manipulator.

An important point to note is that the above methods are subject to local minima and in some cases will fail to converge on the solution. The choice of the initial joint configuration $\vec{q}_0$ is important.
An alternative approach is to re-start an IK problem with a new random $\vec{q}_0$ after a few $20 \sim 50$ iterations rather than persist with a single attempt with $500 \sim 5\, 000$ iterations. This is a simple but effective method of performing a global search for the IK solution.

We display a comparison of IK methods presented in this tutorial in Table \ref{tab:ik}.
Table \ref{tab:ik} shows results for several IK algorithms trying to solve for $10\, 000$ randomly generated reachable end-effector poses using a six degree-of-freedom UR5 manipulator. We show each method initialised with a random valid $\vec{q}_0$ and a maximum of 500 iterations to reach the goal before being declared infeasible. We also show each of the methods with 100 searches to reach the goal where a new search is initialised with a new random valid $\vec{q}_0$ after 30 iterations. The iterations recorded for this method include the iterations from failed searches.

We have presented some useful IK methods but it is by no means an exhaustive list. Sugihara \cite{ik3} provides a good comparison of many different numerical IK methods. In Part 2 of this tutorial, we explore advanced IK techniques which incorporate additional constraints into the optimisation.


\subsection*{Manipulator Performance Metrics}

Manipulator performance metrics seek to quantify the performance of a manipulator in a given configuration. In this section, we explore two common manipulator performance metrics based on the manipulator Jacobian. A full survey of performance metrics can be found in \cite{manip2}. It is important to note several considerations on how manipulator-based performance metrics should be used in practice. Firstly, the metrics are unitless, and the upper bound of a metric depends on the manipulator kinematic model (i.e. joint types and link lengths) and the units chosen to represent translation and orientation. Consequently, metrics computed for different manipulators are not directly comparable. Secondly, the manipulator Jacobian contains three rows corresponding to translational rates, and three rows corresponding to angular rates. Therefore, any metric using the whole Jacobian will produce a non-homogeneous result due to the mixed units. Depending on the manipulator scale, this can cause either the translational or rotational component to dominate the result. This problem also arises in manipulators with mixed prismatic and revolute joints. In general, the most intuitive use of performance metrics comes from using only the translational or rotational rows of the manipulator Jacobian (where the choice of which depends on the use case), and only using the metric on a manipulator comprising a single joint type \cite{manip2}.

\paragraph{Manipulability Index} The Yoshikawa manipulability index \cite{manip} is the most widely used and accepted performance metric \cite{manip2}. The index is calculated as
\begin{align} \label{eq:manipulability}
    m(\vec{q}) = \sqrt{
        \mbox{det}
        \Big(
            \emat{J}(\vec{q})
            \emat{J}(\vec{q})^\top
        \Big)
    }
\end{align}
where $\emat{J}(\vec{q}) \in \mathbb{R}^{3 \times n}$ is either the translational or rotational rows of $\mat{J}(\vec{q})$ causing  $m(\vec{q})$ to describe the corresponding component of manipulability. 
Note that Yoshikawa used $\mat{J}(\vec{q})$ instead of $\emat{J}(\vec{q})$ in (\ref{eq:manipulability}) but we describe it so due to limitations of Jacobian-based measures previously discussed.
The scalar $m(\vec{q})$ describes the volume of a 3-dimensional ellipsoid -- if this ellipsoid is close to spherical, then the manipulator can achieve any arbitrary end-effector (translational or rotational depending on $\emat{J}(\vec{q})$) velocity. The ellipsoid is described by three radii aligned with its principal axes. A small radius indicates the robot's inability to achieve a velocity in the corresponding direction. At a singularity, the ellipsoid's radius becomes zero along the corresponding axis and the volume becomes zero. If the manipulator's configuration is well conditioned, these ellipsoids will have a larger volume. Therefore, the manipulability index is essentially a measure of how easily a manipulator can achieve an arbitrary velocity.

\paragraph{Condition Number} The condition number of the manipulator Jacobian was proposed as a performance measure in \cite{cond}. The condition number is
\begin{align}
    \kappa =
    \dfrac{\sigma_{\mbox{max}}}
    {\sigma_{\mbox{min}}}
    \in [1, \ \infty ]
\end{align}
where $\sigma_{\mbox{max}}$ and $\sigma_{\mbox{min}}$ are the maximum and minimum singular values of $\emat{J}(\vec{q})$ respectively. The condition number is a measure of velocity isotropy. A condition number close to $1$ means that the manipulator can achieve a velocity in a direction equally as easily as any other direction. However, a high condition number does not guarantee a high manipulability index where the manipulator may struggle to move in all directions.


\section*{Acknowledgments}

We are grateful to the anonymous reviewers whose detailed and insightful comments have improved this article.
This research was conducted by the Australian Research Council project number CE140100016, and supported by the QUT Centre for Robotics (QCR). We would also like to thank the members of QCR who provided valuable feedback and insights while testing this tutorial and associated Jupyter Notebooks.


\section*{Conclusions}

In Part 1 of this tutorial, we have covered foundational aspects of manipulator differential kinematics. We first detailed a procedure for describing the kinematics of any manipulator and used this model to derive formulas for calculating the forward and first-order differential kinematics. We then detailed some applications unlocked by these formulas, including reactive motion control, inverse kinematics and methods which describe the performance of a manipulator at a given configuration. In Part 2, we explore second-order differential kinematics and detail how it can improve applications detailed in Part 1 while also unlocking new applications.

\vspace{-3pt}


{\footnotesize
\printbibliography[title={References}]}

@article{dkt2,
  author={Haviland, Jesse and Corke, Peter},
  title={Manipulator Differential Kinematics: Part 2: Acceleration and Advanced Applications},
  journal={IEEE Robotics \& Automation Magazine}, 
  year={2023},
  pages={2-12},
  doi={10.1109/MRA.2023.3270221}
}

@article{stocco1999use,
  title={On the use of scaling matrices for task-specific robot design},
  author={Stocco, Leo J and Salcudean, Septimiu E and Sassani, Farrokh},
  journal={IEEE Transactions on Robotics and Automation},
  volume={15},
  number={5},
  pages={958--965},
  year={1999},
  publisher={IEEE}
}

@inproceedings{mansard2009versatile,
  title={A versatile generalized inverted kinematics implementation for collaborative working humanoid robots: The stack of tasks},
  author={Mansard, Nicolas and Stasse, Olivier and Evrard, Paul and Kheddar, Abderrahmane},
  booktitle={2009 International conference on advanced robotics},
  pages={1--6},
  year={2009},
  organization={IEEE}
}

@article{vs,
  title={A tutorial on visual servo control},
  author={Hutchinson, Seth and Hager, Gregory D and Corke, Peter I},
  journal={IEEE transactions on robotics and automation},
  volume={12},
  number={5},
  pages={651--670},
  year={1996},
  publisher={IEEE}
}

@article{cond,
  title={Articulated hands: Force control and kinematic issues},
  author={Salisbury, J Kenneth and Craig, John J},
  journal={The International journal of Robotics research},
  volume={1},
  number={1},
  pages={4--17},
  year={1982},
  publisher={Sage Publications Sage UK: London, England}
}

@article{manip2,
  title={Manipulator performance measures-a comprehensive literature survey},
  author={Patel, Sarosh and Sobh, Tarek},
  journal={Journal of Intelligent \& Robotic Systems},
  volume={77},
  number={3},
  pages={547--570},
  year={2015},
  publisher={Springer}
}

@article{manip,
	author = "Yoshikawa, Tsuneo",
	doi = "10.1177/027836498500400201",
	eprint = "https://doi.org/10.1177/027836498500400201",
	journal = "The International Journal of Robotics Research",
	number = "2",
	pages = "3--9",
	title = "{Manipulability of Robotic Mechanisms}",
	volume = "4",
	year = "1985"
}

@inproceedings{chan,
  title={General inverse kinematics with the error damped pseudoinverse},
  author={Chan, Stephen K and Lawrence, Peter D},
  booktitle={Proceedings. 1988 IEEE international conference on robotics and automation},
  pages={834--839},
  year={1988},
  organization={IEEE}
}

@article{wampler,
  title={Manipulator inverse kinematic solutions based on vector formulations and damped least-squares methods},
  author={Wampler, Charles W},
  journal={IEEE Transactions on Systems, Man, and Cybernetics},
  volume={16},
  number={1},
  pages={93--101},
  year={1986},
  publisher={IEEE}
}

@inproceedings{ik4,
  title={Adaptive non-linear least squares for inverse kinematics},
  author={Deo, Arati S and Walker, Ian D},
  booktitle={[1993] Proceedings IEEE International Conference on Robotics and Automation},
  pages={186--193},
  year={1993},
  organization={IEEE}
}

@phdthesis{openrave,
 author = "Rosen Diankov",
 title = "Automated Construction of Robotic Manipulation Programs",
 school = "Carnegie Mellon University, Robotics Institute",
 month = "August",
 year = "2010",
 number= "CMU-RI-TR-10-29",
 url={http://www.programmingvision.com/rosen_diankov_thesis.pdf},
}

@incollection{re0,
  title={Real-time obstacle avoidance for manipulators and mobile robots},
  author={Khatib, Oussama},
  booktitle={Autonomous robot vehicles},
  pages={396--404},
  year={1986},
  publisher={Springer}
}

@inproceedings{pot,
  title={Movement reproduction and obstacle avoidance with dynamic movement primitives and potential fields},
  author={Park, Dae-Hyung and Hoffmann, Heiko and Pastor, Peter and Schaal, Stefan},
  booktitle={IEEE International Conference on Humanoid Robots},
  year={2008},
}

@article{ik3,
  title={Solvability-unconcerned inverse kinematics by the Levenberg--Marquardt method},
  author={Sugihara, Tomomichi},
  journal={IEEE Transactions on Robotics},
  volume={27},
  number={5},
  pages={984--991},
  year={2011},
  publisher={IEEE}
}

@inproceedings{rtb,
  title        = {Not your grandmother's toolbox--the Robotics Toolbox reinvented for {P}ython},
  author       = {Corke, Peter and Haviland, Jesse},
  booktitle    = {2021 IEEE international conference on robotics and automation (ICRA)},
  year         = {2021},
  organization = {IEEE}
}

@article{neo,
  title={{NEO}: A Novel Expeditious Optimisation Algorithm for Reactive Motion Control of Manipulators},
  author={Haviland, Jesse and Corke, Peter},
  journal={IEEE Robotics and Automation Letters},
  year={2021}
}

@ARTICLE{ik1,
  author={ {Pyung Chang}},
  journal={IEEE Journal on Robotics and Automation}, 
  title={A closed-form solution for inverse kinematics of robot manipulators with redundancy}, 
  year={1987},
  volume={3},
  number={5},
  pages={393-403},}

@ARTICLE{rrmc,  
  author={D. E. {Whitney}},  
  journal={IEEE Transactions on Man-Machine Systems},   
  title={Resolved Motion Rate Control of Manipulators and Human Prostheses},   
  year={1969},  
  volume={10},  
  number={2},  
  pages={47-53},
}

@ARTICLE{mmc2,
author={D. {Guo} and Y. {Zhang}},
journal={IEEE Transactions on Industrial Electronics},
title={Acceleration-Level Inequality-Based {MAN} Scheme for Obstacle Avoidance of Redundant Robot Manipulators},
year={2014},
volume={61},
number={12},
pages={6903-6914},
doi={10.1109/TIE.2014.2331036},
ISSN={1557-9948},
month={Dec},}

@article{mmc,
    title={Maximising Manipulability During Resolved-Rate Motion Control},
    author={Jesse Haviland and Peter Corke},
    year={2020},
    journal={arXiv preprint arXiv:2002.11901},
}

@book{peter,
  doi = {10.1007/978-3-031-06468-5},
  year = {2023},
  publisher = {Springer Nature},
  author = {Peter Corke},
  title = {Robotics,  Vision and Control: fundamental algorithms in Python},
  edition = 3
}

@article{ets,
  title={A simple and systematic approach to assigning {D}enavit--{H}artenberg parameters},
  author={Corke, Peter I},
  journal={IEEE transactions on robotics},
  volume={23},
  number={3},
  pages={590--594},
  year={2007},
  publisher={IEEE}
}

@article{dh,
  title={A kinematic notation for lower pair mechanisms based on matrices},
  author={Hartenberg, Richard S and Denavit, Jacques},
  year={1955}
}


\end{document}